\documentclass[conference]{IEEEtran}
\IEEEoverridecommandlockouts
\usepackage{cite}
\usepackage{amsmath,amssymb,amsfonts}
\usepackage{algorithmic}
\usepackage{graphicx}
\usepackage{textcomp}
\usepackage{float}
\usepackage{xcolor}
\def\BibTeX{{\rm B\kern-.05em{\sc i\kern-.025em b}\kern-.08em
    T\kern-.1667em\lower.7ex\hbox{E}\kern-.125emX}}
\begin{document}

\title{Detecting Korean Food Using Image using Hierarchical Model\\
{\footnotesize \textsuperscript{}}
\thanks{}
}

\author{
\IEEEauthorblockN{Hoang Khanh Lam}
\IEEEauthorblockA{\textit{Department of Computer Engineering} \\
\textit{Dong-A University}\\
Busan, South Korea  \\
hoangkhanh9119@gmail.com}
\and
\IEEEauthorblockN{ Kahandakanaththage Maduni Pramuditha Perera}
\IEEEauthorblockA{\textit{Department of Computer Engineering} \\
\textit{Dong-A University }\\
Busan, South Korea \\
madunipperera6@gmail.com}
}

\maketitle

\begin{abstract}
A solution was made available for Korean Food lovers who have dietary restrictions to identify the Korean food before consuming. Just by uploading a clear photo of the dish, people can get to know what they are eating. Image processing techniques together with machine learning helped to come up with this solution. 
\end{abstract}


\section{Introduction}
With the immense popularity towards Korean culture, Korean food has gained huge recognition and popularity throughout the world. Most of the foreign tourists come to South Korea to enjoy the Korean food culture. On the other hand, since most of them do not speak or understand the Korean language, selecting food items suitable for them has become a crisis.   

Also nowadays, health problems related to eating and food in general are increasing in both quantity and danger. Especially, most of the foreigners living in South Korea have dietary restrictions due to religion, health-related issues or personal preferences. Besides, being allergic to accidentally eating unfamiliar ingredients in new dishes, especially when traveling, can easily lead to unpredictable consequences. 
Food Image Recognition involves the utilization of visual data to determine food name/label. It has the potential to contribute to various applications, such as dietary assessment, food inspection, food recognition, and food recommendation.

\section{Challenges}
Obtaining a suitable data-set was the very first challenge faced. After obtaining a proper data-set the model training phase was initiated. Similarity among different types of Korean Food made the training phase more of a challenge. Differentiating food with similar color and texture was difficult. At the same time dealing with poorly taken pictures and incomplete pictures was also an obstacle. Since most of the Korean dishes come with several side dishes, detecting multiple foods at the same time was also an added difficulty.  

\section{Related Work}
Research in the field of computer vision and food image recognition, including the identification of specific cuisines such as Korean food, is continuously evolving. Back in 2019, Korea Food Research Institute, Wanju conducted a research on "The development of food image detection and recognition model of Korean food for mobile dietary management". According to the results of this study, higher accuracy level of 91.3\% was obtained with a 0.4ms recognition time. In this study, they have collected food images by taking pictures or by searching web images and built an image dataset for use in training a complex recognition model for Korean food. Augmentation techniques were performed in order to increase the dataset size\cite{doi:10.4162/nrp.2019.13.6.521}

On October 2022, group of researches from Department of Computer Science and Engineering, University of Seoul have conducted a study on "Development of Korean Food Image Classification Model Using Public Food Image Dataset and Deep Learning Methods". Their model can be used in a system that automatically classifies the type of food that the user has consumed. This work will benefit the community of users and researchers using trained models. It also benefits users of systems that automatically classify and record the types of food they take. Users can classify images of similar Korean food into 150 classes using the model and record what
they eat. The key contribution of this study to the research community is that they created a pre-processed dataset for training a Korean food image classification model. Also,they have used several pre-trained deep neural networks to train
models that classify Korean food images into 150 classes and evaluated the classification accuracy and time required for model training\cite{67568}.

\section{Dataset}
%Write about the expectation we want the data have to (taken from phone, various,  label in orginal Korean name, not English name, ...)
%Explain the reason for each.
The data uploaded for detection have to be very clear. At the same time, the photos should be taken from a mobile phone as the images used for training and deploying the system were also obtained through the camera of the mobile phone. Since mobile phones are portable, it is easy for users to capture a wide range of food images.This ensures a large and diverse data-set for training food image recognition models.  
Mobile phone photos are captured in real-world scenarios with different lighting, angles, and backgrounds. This variety aids in training models to identify food in diverse environments, enhancing their practical usability. Using mobile phone photos for food recognition is convenient for users, as they can easily snap pictures of their meals without the need for specialized equipment. This ease of use encourages greater user participation and contributes to the creation of extensive data-sets.
% 
%...........................................

\subsection{Public Data}\label{AA}
%Write about the way we search for data set (aihub,..roboflow ):difficulties on aihub and details info about the roboflow data
Once done with the background research, obtaining a proper dataset was a huge crisis. Several options as aihub, Roboflow were considered. Though there were several suitable datasets on aihub, obtaining such data was difficult for foreigners. After revising many datasets, one suitable for our system was obtained from Roboflow. 
\subsection{Data-set Construction}
%Write about the way we cut off the food and put random together on random background to augment the data set.
As a basic approach, categorizing food based on visual aspects was initiated. A mechanism to group food based on color, texture, contrast, size, shape, arrangement, temperature, layering and theme was taken into account. Korean food were divided into four main categories namely, Main Dish, Rice, Soup and Side Dish. Each category was divided further into sub categories accordingly. After completing the Background Research, application of a hierarchical model to group data was considered. 
\begin{figure}[H]
    \centering
    \includegraphics[scale=0.3]{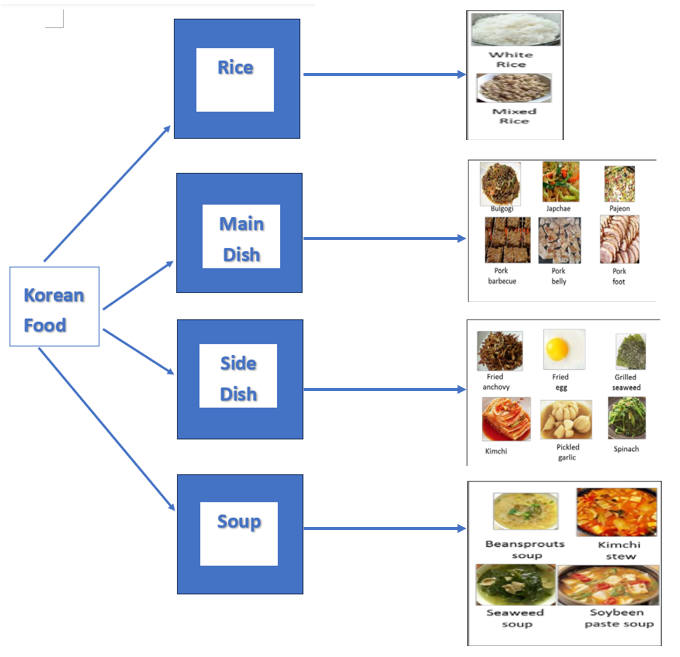}
    \caption{Caption}
    \label{fig:enter-label}
\end{figure}

To create the Korean food recognition system, following steps were followed.  Defining the scope of the dataset, compiling a list of dishes, gathering high-quality images, ensuring diversity, and using data augmentation techniques. Labeling each image with the corresponding dish name, organizing the dataset into training, validation, and testing sets, and balancing the dataset. Applicable additional metadata was included. Documentation that describes dataset was created, including the number of images, dish categories, and relevant details. The system was trained and evaluated using machine learning or deep learning techniques.

\subsection{Hierarchical Dataset}
%Write about why we want to create hierarchical dataset, why this method suitable for food recoginition. 
A hierarchical model is a structure that organizes elements into levels or layers, where each level represents a specific degree of abstraction or detail. The arrangement of elements reflects their hierarchical relationships and dependencies. Hierarchical models are often represented as trees, where the topmost level is the root, and subsequent levels branch out like the limbs of a tree. This visual representation makes it easy to understand the relationships and navigate through the structure.

A hierarchical model could be structured based on ingredients, allowing for the categorization of dishes according to shared components. For example, dishes made with similar base ingredients like kimchi, red chili paste, or rice could be grouped together. Also can provide a user-friendly interface, allowing users to navigate through categories and subcategories easily. In the context of designing a menu, a hierarchical model can assist in creating a well-structured and visually appealing layout. Dishes can be grouped logically, making it easier for customers to navigate and choose their preferences. 

\section{Our Method}
In this part, we present a system for categorizing food products. The approach attempts to address two main problems: the dataset's class imbalance and the absence of obvious visual relationships between various food products. In order to overcome these difficulties, we establish a tier of hierarchy between "food types" and "food items". This is achieved in the training phase by updating the Convolution Neural Network (CNN) model by multi-stage transfer learning and clustering visually related food items iteratively. The trained model in the multi-stage transfer learning step is used directly for food image classification during the validation and testing stages. Besides, to prove the effectiveness of the hierarchical model, we also train the flat classification model as a baseline model.
\subsection{Flat Classification}
This research proposes a YOLOv8-based method for smart city food image detection.  The suggested methodology aims to mitigate some constraints of prior research and offer enhanced precision, instantaneous identification, adaptability, diminished false alerts, and economical viability. 

The YOLO collection of algorithms has gained interest in computer vision. Because it maintains a high level of accuracy while maintaining a small model size, YOLO is quite popular. YOLO models may be trained on a single GPU, making them accessible to a broad range of developers. It may be reasonably installed by machine learning specialists on edge hardware or in the cloud. YOLOv8, the most advanced and recent YOLO technique, can be used for segmentation, object recognition, and image classification. Ultralytics, the same company that created the influential YOLOv5 model that helped to define the industry, is the producer of Yolo v8. YOLOv8 contains a few architectural improvements and updates over YOLOv5\cite{a2023_yolov8}.

The YOLOv8 model does not use anchors. It suggests that the object's center is explicitly estimated rather than the item's distance from a known anchor box. After inference, a difficult post-processing step called Non-Maximum Suppression (NMS) is sped up by anchorfree detection, which reduces the quantity of box predictions. Five models are available for identification, segmentation, and classification (YOLOv8n, YOLOv8s, YOLOv8m, YOLOv8l, and YOLOv8x, respectively). YOLOv8 Nano is the smallest and fastest of them all, whereas YOLOv8x is the most accurate but slowest of them all. The following are the differences from YOLOv5:
\begin{itemize}
    \item C2f module used in place of C3 module.
    \item Change the Backbone’s initial 6x6 Conv to a 3x3 Conv.
    \item Remove Convs Nos. 10 and 14 from the YOLOv5 configuration.
    \item Change the initial 1x1 Conv in the bottleneck to a 3x3 Conv.
    \item Remove the objectness step using the decoupled head.
\end{itemize}

\begin{figure*}
    \centering
    \includegraphics[width=\textwidth,height=18cm]{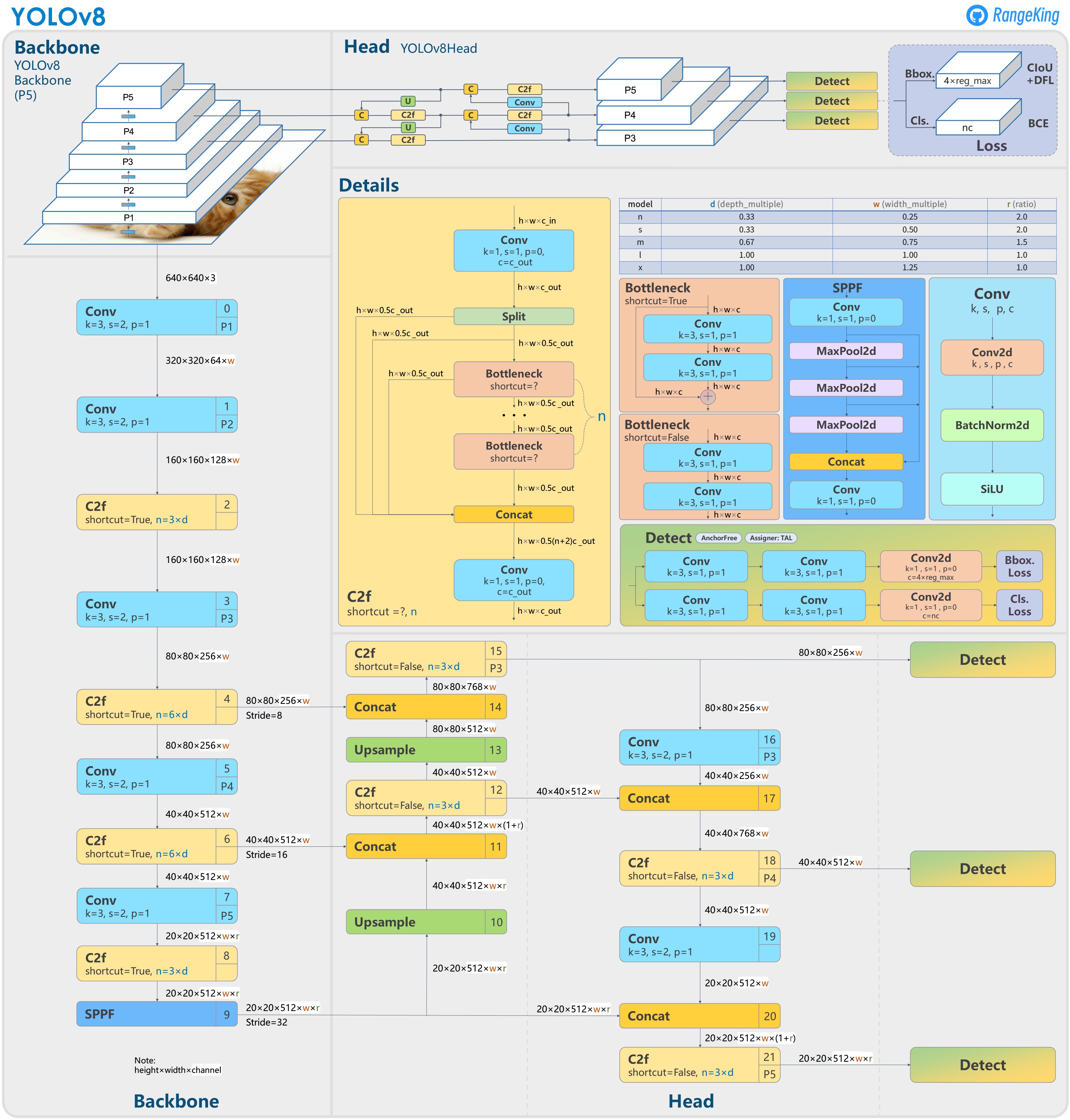}
    \caption{YOLO v8 architecture\cite{_2023_brief}}
    \label{fig:yolo}
\end{figure*}

The fundamental building block was changed, using C2f in place of C3, and the stem's original 6x6 conv was swapped out for a 3x3. The module is summarized in Figure \ref{fig:yolo}, where "f" is the total number of features, "e" the growth rate, and CBS is a block consisting of a Conv, a BatchNorm, and a SiLU thereafter. The kernel dimension of the first convolution was changed from 1x1 to 3x3, however the bottleneck remains the same as in YOLOv5.

%...........................................
\subsection{Hierarchical Classification}
We use multi-stage hierarchical transfer learning to take advantage of the hierarchical structure that has been built and to address the problem of class imbalance in food products. By using this method, we can move knowledge from the top level—food types—to the lower level—food items. The pipeline of our multi-stage hierarchical transfer learning approach is depicted in . Our training procedure is iterative and consists of two stages.
\begin{itemize}
    \item First stage: We use a CNN model (Yolov8) that has been pre-trained on ImageNet for the first iteration [8]. Next, using food kinds as labels, this model is connected to a fully connected layer with a size of T = 4. We keep the parameters in the core of the CNN model that was trained in the previous iteration for use in later iterations. Using food kinds as labels, a new CNN model is built from scratch and connected to a fully connected layer with a size of T = 4.
    \item Second stage: We build a new CNN model by leveraging the parameters from the foundation of the CNN model trained in the second stage. This model is trained with food items as labels and is linked to a fully connected layer with a size of N=32.
\end{itemize}
We use the cross-entropy loss, which is expressed as follows, as our loss function at each training phase:
\begin{equation}
    L=-\sum_{i=1}^{N}y_{s,i}log(p_{s,i})
\end{equation}
Here, \(s\) denotes the stage at which the model is being trained. \(y_{s,i}\) represents the ground truth label \(i\) at stage \(s\) , and  indicates the confidence score for predicting label \(i\) at stage \(s\).

The CNN model is then used in the food item merging process to produce new merging results after being improved through this multi-stage hierarchical transfer learning process. The next iteration of the multi-stage hierarchical transfer learning process is then started by transferring the parameters from the model's core. This loop keeps on until either the risk of over fitting is mitigated by reaching a maximum of 5 iterations, or the validation loss on food item classification stops decreasing in subsequent rounds. Ultimately, predictions about food items are retrieved using the CNN model that was trained in the last iteration of the second stage. The preserved model from the first iteration's stage is utilized to forecast food categories for experimental comparison with related research.
\begin{figure}[H]
    \centering
    \includegraphics[scale = 0.3]{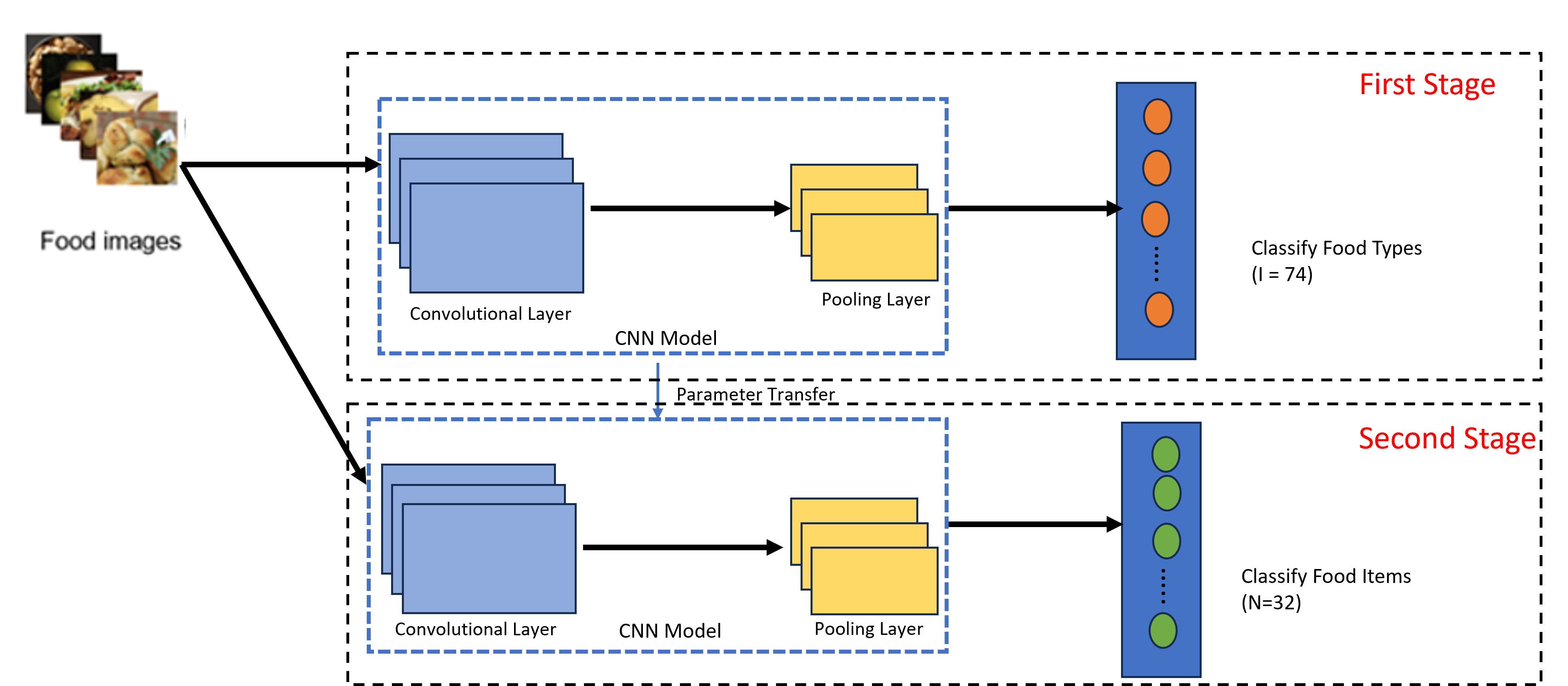}
    \caption{Overall Hierarchical Classify Model}
\end{figure}

%...........................................
\section{Experiment}
% ...........................................
We evaluate our proposed method based on average classification accuracy of predicting food items. We divide the Food Image dataset using a 8:1:1 ratio into training, validation, and testing sets in order to train the CNN model. The validation set is used to evaluate whether the trained model should be saved for further testing, whereas the training set is used to train the model. The testing set is set aside for the last assessment of the model.
The CNN architecture of our suggested solution is based on the Yolov8. We employ the Adam optimizer for optimization, starting with a 0.001 learning rate. The multi-stage hierarchical transfer learning process is taught over 100 epochs for each stage. If the loss does not decrease after each iteration, we cut the initial learning rate by a factor of 0.03. 
\begin{table} [H]
\centering
\begin{tabular}{|c||c|}
    \hline
    Methods & Average accuracy (\%) \\
    \hline
    Flat Classification & 85.32 \\
    Hierarchical Classification & 88.50 \\
    \hline
\end{tabular}

\caption{Average Accuracy in our experiment}
\end{table}
At first, the flat classification model was used to train the model, and an accuracy of 85.32\% was obtained. Since the data is not arranged and sorted in a proper manner,  decided to come up with a more neat and organized classification using a Hierarchical model. 

The hierarchical approach makes it easier to interpret and explain the relationships between different classes. This can be important in applications where understanding the reasoning behind the classification is crucial. The hierarchical approach can reduce the complexity of the classification task by breaking it down into a series of simpler subtasks. Each level of the hierarchy deals with a more specific aspect of the classification, making it easier to manage and interpret. After training the model using the Hierarchical approach, an accuracy of 88.50\% was achieved and this outperformed the flat classification. 
\section{Conclusion}
% ...........................................
In this work, we took advantage of a Korean food image dataset, which assigns related food item to each food image. Within the dataset, we treat food items as subclasses of food kinds, creating a hierarchical structure. We can obtain the relevant nutritional composition data by predicting food items, which advances the objective of image-based dietary evaluation. We used a multi-stage hierarchical transfer learning approach to update the CNN model for extracting image features iteratively during the training phase in order to create visual relations among food items. This approach can also address the problem of class imbalance across food items.

Although these efforts, there exist additional plausible approaches that may enhance the precision of food item classification even further. Our goal is to use concepts from multi-modal learning to our future work to improve the way we classify food products and, consequently, the way food image categorization on food items is refined.

\bibliographystyle{IEEEtran}
\bibliography{ref}

\vspace{12pt}

\end{document}